\documentclass[sigconf,nonacm]{acmart}

\AtBeginDocument{%
  }

\usepackage{algorithm}
\usepackage{algpseudocode}
\usepackage{amsmath}
\usepackage{amssymb}
\usepackage{xcolor}
\usepackage{booktabs}
\usepackage{multirow}
\usepackage{array}
\usepackage{colortbl}
\usepackage{placeins}
\usepackage{graphicx}  
\usepackage{url}       
\usepackage{listings}
\begin{document}

\title{FLARE: Few-shot Learning-based Adaptive Reflective Engine}

\author{Dhanasekar Sundararaman, Bharat Gandhi, Aashna Garg, Minjie Li}
\affiliation{%
\institution{Microsoft}
\city{Redmond}
\state{WA}
\country{USA}}
\email{{dhanasekars, aashnagarg}@microsoft.com} 
\renewcommand{\shortauthors}{Sundararaman et al.}

\begin{abstract}
  Large language models (LLMs) are increasingly deployed in complex, compound AI systems where performance hinges on the quality of prompts. Recent state-of-the-art optimizers like GEPA (Genetic-Pareto) have argued that reflective instruction evolution can outperform traditional reinforcement learning and few-shot optimization. In this work, we challenge this shift by introducing FLARE (Few-shot Learning-based Adaptive Reflective Engine), a framework that leverages advanced reflective mechanisms and a small set of few-shot reference examples to optimize instructions. We evaluate our method across a diverse suite of benchmarks—spanning retrieval-augmented reasoning (HotPotQA, MedQA, 2WikiMultiHopQA), tool calling, and multi-label emotion classification (GoEmotions)—using the GPT-5 series of models. Our results demonstrate that FLARE consistently outperforms GEPA, winning on every task--model pair: it achieves gains of up to +14.2 points on HotPotQA (52.2 vs.\ GEPA's 42.2 with GPT-5-Chat), reaches 87.0\% on tool calling (vs.\ 81.0\% for GEPA), and lifts GoEmotions micro-F1 to 52.7\% (+15.3) with GPT-5.1 on the full 5{,}408-example test split, more than doubling GEPA's +5.7 gain. Beyond raw accuracy, FLARE is also strikingly data-efficient: on GoEmotions it reaches its peak performance using as few as 100 validation examples, while remaining markedly more stable across random seeds than GEPA. Our findings suggest that while reflective instructions are powerful, the strategic optimization of few-shot learning remains a critical frontier for maximizing the potential of next-generation LLMs.
\end{abstract}


\maketitle

\section{Introduction}
\label{sec:intro}
The advent of Frontier Large Language Models (LLMs) has shifted the challenge of AI development from model training to the engineering of complex, compound AI systems. In these systems, performance is governed by the quality of prompts, which must orchestrate reasoning, tool manipulation, and domain-specific logic. However, manual prompt refinement is unscalable and often fails to capture the subtle nuances required for peak performance. Automated prompt optimization has emerged as a solution, moving from simple discrete searches to sophisticated reflective frameworks.

A prominent recent advancement is \textit{Genetic-Pareto} (GEPA) \cite{Agrawal2025GEPA}, which utilizes natural language reflection to evolve high-level instructions. GEPA posits a significant shift in the field: as models become more capable, instruction-only evolution can outperform traditional few-shot optimization. While GEPA provides a powerful mechanism for rule discovery, we argue that instruction-only reflection often lacks the \textit{grounding} necessary for tasks where the delta between success and failure lies in subtle execution patterns that abstract rules cannot fully capture.

In this work, we introduce FLARE (Few-shot Learning-based Adaptive Reflective Engine),\footnote{Official implementation: \url{https://github.com/microsoft/FLARE---Few-shot-Learning-based-Adaptive-Reflective-Engine}} a framework that grounds reflective prompt optimization in explicit, per-instance failure signals. Whereas GEPA reflects on execution traces to evolve high-level instructions and searches a candidate pool via genetic--Pareto selection, FLARE uses the reasoning capabilities of a frontier GPT-5-series model to diagnose \emph{concrete} mistakes and repair them directly.

In our fine-grained error-aware feedback loop, a frontier GPT-5-series model performs a specialized "thinking" phase over each validation example alongside the primary model's prediction and an explicit correctness label (\texttt{Match: WRONG}). Rather than evolving abstract rules, it autonomously diagnoses the root cause of each failure and rewrites the prompt with targeted corrections, turning every observed mistake into a direct, surgical instruction update. This lets the model adapt its predictive logic based on a grounded, model-driven analysis of its own shortcomings.

Our empirical evaluation on the GPT-5 series demonstrates that this iterative, error-focused grounding significantly outperforms GEPA's instruction-only evolution across a wide array of benchmarks:

\begin{itemize}
    \item \textbf{Data-Efficient Classification:} On multi-label emotion classification (\texttt{GoEmotions}) with GPT-5.1, FLARE lifts micro-F1 to \textbf{52.7\%}, a \textbf{+15.3} improvement over the baseline that more than doubles GEPA's \textbf{+5.7} gain---while reaching its peak with as few as 100 validation examples.
    \item \textbf{Superior Reasoning Capabilities:} On \texttt{HotPotQA} with GPT-5-Chat, FLARE achieves an optimized mean of \textbf{52.2}, a \textbf{+14.2} improvement over the baseline and a \textbf{10-point lead} over GEPA (\textbf{42.2}).
    \item \textbf{Gains in Complex Tool Usage:} For tool-calling tasks, our framework achieves \textbf{87.0\%} accuracy, compared to the \textbf{81.0\%} ceiling reached by GEPA.
\end{itemize}

The core contribution of this paper is the demonstration that prompt optimization is most effective when reflection is grounded in the iterative diagnosis of concrete failure modes. By coupling error-aware feedback with a small set of few-shot reference examples, we show that the recently argued "shift" away from few-shot grounded optimization is premature. In its place, we offer a state-of-the-art methodology for maximizing the utility of today's most capable models.

\section{Related Work}

Recent advances in automated prompt optimization have demonstrated significant improvements in LLM performance across various tasks. DSPy~\cite{Khattab2024DSPy} introduces a declarative framework that compiles language model calls into self-improving pipelines, enabling systematic optimization of prompts through programmatic composition. Building on this foundation, GEPA~\cite{Agrawal2025GEPA} demonstrates that reflective prompt evolution can outperform traditional reinforcement learning approaches by iteratively refining prompts based on performance feedback. Yang et al.~\cite{Yang2024OPRO} propose OPRO, which frames prompt optimization as an optimization problem where LLMs themselves serve as optimizers, using meta-prompting to generate and refine task-specific instructions on benchmarks such as GSM8K. Within the DSPy ecosystem, the BootstrapFewShot optimizer~\cite{Khattab2024DSPy} synthesizes few-shot demonstrations from labeled examples, while MIPROv2~\cite{Khattab2023OptimizingInstructions} jointly optimizes instructions and demonstrations through Bayesian search over candidate proposals; GEPA reports surpassing MIPROv2, which motivates our choice of GEPA as the primary optimization baseline.

Beyond these general-purpose methods, optimizing prompts for RAG systems presents unique challenges due to the interaction between retrieval and generation components. Rodrigues and Branco~\cite{RodriguesBranco2024MetaPromptingRAG} apply meta-prompting techniques to RAG pipelines to improve end-to-end performance, and Opsahl-Ong et al.~\cite{Khattab2023OptimizingInstructions} optimize instructions and demonstrations for multi-stage language model programs, achieving strong results on multi-hop question answering datasets like HotPotQA~\cite{Yang2018HotpotQA}. RAG-Gym~\cite{xiong2025rag} instead supervises the search process of retrieval agents, evaluating performance on multi-hop datasets including HotPotQA and 2WikiMultiHopQA~\cite{Ho2019TwoWikiMultiHopQA}. Complementary approaches enhance RAG through architectural innovations rather than prompt design: IRCoT~\cite{Trivedi2022IRCoT} interleaves retrieval with chain-of-thought reasoning, Self-RAG~\cite{Asai2024SelfRAG} learns when to retrieve and critically evaluate retrieved content, Active RAG~\cite{Jiang2023ActiveRAG} determines when retrieval is necessary during generation, and LongRAG~\cite{Jiang2024LongRAG} leverages long-context LLMs to process more retrieved documents simultaneously, while Park et al.~\cite{Park2024EmulatingRAG} emulate RAG behavior through prompt engineering without an explicit retriever. More general frameworks such as Promptomatix~\cite{Murthy2025Promptomatix}---which we adopt as a baseline---and APE~\cite{Zhou2023HumanLevelPromptEngineers} automatically search the prompt space to synthesize high-performing prompts across tasks. Finally, for classification, Pryzant et al.~\cite{Pryzant2023AutomaticPromptOptimization} propose automatic prompt optimization using gradient descent and beam search, and Lieander et al.~\cite{Lieander2025TwoGradientPromptOpt} introduce a two-gradient approach that combines multiple optimization signals to improve LLM classification. Beyond the research literature, commercial providers also ship prompt-optimization tooling; we additionally compare against OpenAI's Prompt Optimizer,\footnote{\url{https://developers.openai.com/api/docs/guides/prompt-optimizer}} a dashboard tool that rewrites a user-supplied prompt according to current best practices, optionally guided by dataset annotations and grader results, which we include as a widely used practitioner baseline (denoted \textbf{OpenAI} in our tables).

\begin{algorithm}[t]
\caption{FLARE}
\label{alg:custom_opt}
\begin{algorithmic}[1]
\Require Training set $\mathcal{D}_{tr}$, validation set $\mathcal{D}_{val}$, test set $\mathcal{D}_{te}$, metric $m(\cdot)$, iterations $N$
\Ensure Best-performing prompt $p^\star$

\State Initialize LM $\mathcal{L}$ ($\tau = 1.0$, $k_{\max} = 16{,}000$); load system instruction $s_{\text{sys}}$
\State Convert datasets: $\mathcal{T} \leftarrow \phi(\mathcal{D}_{tr})$, $\mathcal{V} \leftarrow \phi(\mathcal{D}_{val})$, $\mathcal{E} \leftarrow \phi(\mathcal{D}_{te})$
\State Extract $p_0 \leftarrow \mathrm{Instr}(P_0)$; evaluate $s^{\text{val}}_0 \leftarrow m(p_0; \mathcal{V})$, $s^{\text{te}}_0 \leftarrow m(p_0; \mathcal{E})$
\State Initialize $p^\star \leftarrow p_0$, $\; s^{\text{val}}_{\text{best}} \leftarrow s^{\text{val}}_0$, $\; H \leftarrow \emptyset$ (window size $\ell = 3$)
\For{$i = 1$ \textbf{to} $N$}
    \State $p \leftarrow \mathcal{O}(p^\star, \mathcal{T}, \mathcal{V}, s_{\text{sys}}, s^{\text{val}}_{\text{best}}, H)$ \Comment{complete train set $\mathcal{T}$, not a sample}
    \State Evaluate $s^{\text{val}} \leftarrow m(p; \mathcal{V})$, $\; s^{\text{te}} \leftarrow m(p; \mathcal{E})$
    \State Append $s^{\text{val}}, s^{\text{te}}$ to $\mathcal{R}_{\text{val}}, \mathcal{R}_{\text{te}}$ \Comment{test tracked for reporting only}
    \State $H \leftarrow \text{last}_\ell\bigl(H \cup \{(s^{\text{val}}, \psi(p))\}\bigr)$
    \If{$s^{\text{val}} > s^{\text{val}}_{\text{best}}$}
        \State $s^{\text{val}}_{\text{best}} \leftarrow s^{\text{val}}$, $\; p^\star \leftarrow p$ \Comment{select by validation}
    \EndIf
\EndFor
\State \Return $p^\star$ \Comment{validation-maximizing prompt, not the final iteration}
\end{algorithmic}
\end{algorithm}

\section{The FLARE Algorithm}

FLARE implements an iterative, error-driven approach to prompt optimization using Azure OpenAI. Unlike GEPA's population-based Pareto frontier evolution, OpenAI's single-pass best-practices rewriting, or Promptomatix's synthetic-data-driven pipeline, our approach operates as a direct feedback loop where an LLM meta-optimizer $\mathcal{O}$ continuously refines a prompt $p$ against task-specific validation performance.  This architecture enables the optimizer to learn from concrete failure patterns rather than relying on population diversity or synthetic demonstrations, resulting in targeted improvements that directly address observed weaknesses in model behavior.

The process begins by initializing language model $\mathcal{L}$ with fixed decoding parameters (temperature $\tau = 1.0$, max completion tokens $k_{\max} = 16{,}000$) and converting the training, validation, and test sets into model-compatible formats $\mathcal{T} \leftarrow \phi(\mathcal{D}_{tr})$, $\mathcal{V} \leftarrow \phi(\mathcal{D}_{val})$, and $\mathcal{E} \leftarrow \phi(\mathcal{D}_{te})$, where transformation $\phi$ constructs DSPy Example objects with appropriate input-output field mappings. A base prompt $p_0 \leftarrow \mathrm{Instr}(P_0)$ is extracted from the task-specific signature and evaluated via metric $m(\cdot)$ to establish baseline scores $s^{\text{val}}_0$ and $s^{\text{te}}_0$. This initialization phase also involves computing predictions $\hat{y}_i \leftarrow \mathcal{L}(p_0, x_i)$ for all validation examples, creating an initial error profile that informs the first optimization step.

Over $N$ optimization iterations, the optimizer updates the prompt through a structured refinement process. At iteration $t$, the meta-optimizer receives the current best-performing prompt $p_{t-1}^\star$ along with detailed feedback constructed from validation predictions. Specifically, for each validation example $(x_i, y_i) \in \mathcal{V}$, we compute the prediction $\hat{y}_i \leftarrow \mathcal{L}(p_{t-1}^\star, x_i)$ and construct an error signal
\[
    e_i = \begin{cases}
        \texttt{CORRECT} & \text{if } m(\hat{y}_i, y_i) = 1 \\
        \texttt{WRONG}(y_i, \hat{y}_i) & \text{otherwise}
    \end{cases}
\]
that includes both the ground truth $y_i$ and the model's incorrect prediction $\hat{y}_i$. The optimizer then generates an improved prompt as
\[
    p_t \leftarrow \mathcal{O}(p_{t-1}^\star,\, \mathcal{T},\, \{(x_i, y_i, e_i)\}_{i=1}^{|\mathcal{V}|},\, s_{\text{sys}},\, c_t,\, H_t),
\]
where $\mathcal{T}$ is the \emph{complete} training set provided for context, $s_{\text{sys}}$ encodes task-specific optimization objectives (e.g., ``maximize F1-score for multi-label classification''), $c_t = s^{\text{val}}_{t-1}$ is the current validation score, and $H_t = \{(s^{\text{val}}_j, \psi(p_j))\}_{j=\max(0,t-3)}^{t-1}$ maintains a sliding window of the $\ell=3$ most recent iterations. The meta-optimizer operates with temperature $\tau_{\mathcal{O}} = 1.0$ (matching the evaluation model) and max tokens $k_{\mathcal{O}} = 16{,}000$, retaining stochasticity to encourage exploration of the prompt space. Notably, both the meta-optimizer and evaluation model use the same stochastic decoding parameters, rather than employing deterministic inference for evaluation.

Each iteration re-evaluates the updated prompt on both validation and test sets to produce new scores $s^{\text{val}}_t$ and $s^{\text{te}}_t$. These scores are appended to result lists $\mathcal{R}_{\text{val}} = [s^{\text{val}}_0, \ldots, s^{\text{val}}_t]$ and $\mathcal{R}_{\text{te}} = [s^{\text{te}}_0, \ldots, s^{\text{te}}_t]$, enabling longitudinal performance tracking. To select the best-performing prompt, we track $p^\star = \arg\max_{p_j} s^{\text{val}}_j$ based on \emph{validation performance} throughout optimization; the held-out test set is used solely for reporting and never informs prompt selection. The optimization runs for $N$ iterations and returns the validation-maximizing prompt $p^\star$.

The orchestration layer supports multiple task types---classification, RAG, and tool calling---with specialized DSPy signatures and conversion functions $\phi_{\text{task}}$. For classification tasks, $\phi$ maps raw labels to sets of emotion IDs; for tool calling, it parses conversational histories and expected tool sequences. In the tool calling domain, the conversion process is particularly nuanced: when ground truth contains multi-tool sequences $[t_1, \ldots, t_k]$, we generate $k$ separate training examples, each focusing on a single tool call $t_j$ with its specific arguments, thereby providing the optimizer with fine-grained feedback on individual tool selection and parameter extraction decisions. This decomposition enables the meta-optimizer to identify whether errors stem from incorrect tool selection, malformed argument values, or improper JSON formatting. The complete training set $\mathcal{T}$ is provided to the meta-optimizer at each iteration, ensuring full visibility into available training patterns without sampling-induced variance.

The longitudinal history $H_t$ provides the meta-optimizer with critical performance context across iterations, enabling it to identify whether prompt modifications are producing monotonic improvements, oscillating around a local optimum, or exploring entirely new solution directions. By exposing both score trajectories and prompt evolution patterns, the history mechanism allows $\mathcal{O}$ to make increasingly informed refinements---for instance, detecting when repeated attempts to improve a specific error pattern are counterproductive and pivoting to alternative reformulation strategies. Furthermore, system instructions $s_{\text{sys}}$ are task-dependent and encode domain-specific constraints such as ``output must be valid JSON matching the tool schema'' for tool calling, guiding the meta-optimizer toward solutions that satisfy both accuracy and format requirements. The stochastic decoding parameters ($\tau = 1.0$ for both optimization and evaluation) introduce variability across runs, enabling more diverse exploration of the prompt space during the meta-optimization process.

\section{Experimental Setup}

\subsection{Datasets}

We evaluate the FLARE approach across three diverse tasks, each presenting unique challenges for language models:

\begin{enumerate}
    \item \textbf{Classification (Emotion Detection)}: Multi-label emotion classification from the GoEmotions dataset~\cite{Demszky2020GoEmotions}. The task requires identifying multiple emotions from 28 possible emotion labels in text snippets. We evaluate classification on the \emph{full} held-out test split (5{,}408 examples) and, holding the reference exemplar set fixed at 50 examples, sweep the validation-set size from 10 to 1{,}000 to study data efficiency (Table~\ref{tab:results_classification_ablation}). Here the reference exemplars are the labeled examples shown to the optimizer as in-context references, while the validation set is what drives the error-based optimization signal.
    
    \item \textbf{Tool Calling}: Function calling and argument extraction tasks drawn from the $\tau^2$-bench benchmark~\cite{Barres2025Tau2Bench} (airline and retail customer-service domains), reformulated as a single-turn tool-selection task. Each example provides a conversational context from which the model must select the appropriate tool and extract correct parameters from natural language.
    
    \item \textbf{Retrieval-Augmented Generation (RAG)}: Question-answering tasks from HotPotQA~\cite{Yang2018HotpotQA}, MedQA~\cite{Jin2020MedQA}, and 2WikiMultiHopQA~\cite{Ho2019TwoWikiMultiHopQA} datasets requiring reasoning over retrieved documents, with multi-hop reasoning in the case of HotPotQA and 2WikiMultiHopQA. The dataset includes complex queries that necessitate synthesizing information from multiple retrieved passages to generate accurate answers.
\end{enumerate}

\subsection{Evaluation Metrics}

We employ task-specific evaluation metrics aligned with the Azure AI Evaluation framework:

\begin{itemize}
    \item \textbf{Classification}: F1-score to handle the multi-label nature of emotion detection
    \item \textbf{Tool Calling}: Combined metric averaging three sub-metrics:
    \begin{enumerate}
        \item Tool Call Accuracy for correct tool selection
        \item Intent Resolution for capturing user intent
        \item Task Adherence for following specifications
    \end{enumerate}
    \item \textbf{RAG}: A retrieval-aware composite scoring factual accuracy against retrieved context, computed as $0.3\times$ retrieval $+\,0.7\times$ generation quality
\end{itemize}

\subsection{Hyperparameters}

 We use Azure OpenAI GPT-5.1 and GPT-5-Chat as the meta-optimizer responsible for analyzing validation performance and rewriting prompts; optimized prompts are then evaluated once on the held-out test dataset. The model operates with temperature=1.0 to encourage diverse prompt variations and escape local optima. The number of optimization iterations was set to 40. Unless otherwise noted, all reported scores are the mean $\pm$ standard deviation over three independent runs (seeds). The complete implementation is built on the DSPy framework\footnote{\url{https://github.com/stanfordnlp/dspy}}, enabling modular prompt templates with automatic signature generation and structured prediction formats.

\section{Results and Discussion}
\label{sec:results}
\begin{table*}
\centering
\caption{Performance comparison of prompt optimization methods on the classification task. For FLARE and GEPA, we report each method's best configuration across the validation-set-size sweep (full sweep in Table~\ref{tab:results_classification_ablation}); GEPA uses its best of the light/heavy budgets. Best optimized score per dataset-LLM combination is shown in \textbf{bold}.}
\label{tab:results_classification}
\begin{tabular}{@{\extracolsep{\fill}}llcccc@{}}
\toprule
\textbf{Dataset} & \textbf{LLM} & \textbf{Optimizer} & \textbf{Unoptimized Mean w/ SD} & \textbf{Optimized Mean w/ SD} & \textbf{$\Delta$} \\
\midrule
\multirow{8}{*}{GoEmotions} & \multirow{4}{*}{GPT-5-Chat} & FLARE & 34.10 $\pm$ 0.12 & \textbf{48.69 $\pm$ 1.80} & +14.6 \\
                           &                               & GEPA & 34.17 $\pm$ 0.31 & 44.87 $\pm$ 4.66 & +10.7 \\
                           &                               & Promptomatix & 36.0 $\pm$ 2.2 & 38.7 $\pm$ 1.2 & +2.7 \\
                           &                               & OpenAI & 36.5 $\pm$ 1.2 & 43.2 $\pm$ 2.2 & +6.7 \\
\cmidrule(lr){2-6}
                           & \multirow{4}{*}{GPT-5.1} & FLARE & 37.46 $\pm$ 0.07 & \textbf{52.72 $\pm$ 0.78} & +15.3 \\
                           &                           & GEPA & 37.96 $\pm$ 0.31 & 43.64 $\pm$ 3.14 & +5.7 \\
                           &                           & Promptomatix & 36.7 $\pm$ 1.7 & 38.0 $\pm$ 0.8 & +1.3 \\
                           &                           & OpenAI & 36.2 $\pm$ 0.7 & 47.8 $\pm$ 1.4 & +11.6 \\
\bottomrule
\end{tabular}
\end{table*}

\begin{table*}
\centering
\caption{Validation-set-size ablation on the classification task. All scores are micro-F1 (\%), mean $\pm$ SD over 3 seeds. Best optimized score per row is shown in \textbf{bold}. GEPA is reported for both the \emph{light} and \emph{heavy} budgets.}
\label{tab:results_classification_ablation}
\begin{tabular}{@{\extracolsep{\fill}}llcccc@{}}
\toprule
\textbf{LLM} & \textbf{Val Size} & \textbf{Baseline} & \textbf{GEPA (light)} & \textbf{GEPA (heavy)} & \textbf{FLARE} \\
\midrule
\multirow{8}{*}{GPT-5.1} & 10 & 36.32 $\pm$ 0.15 & 42.11 $\pm$ 5.16 & 42.01 $\pm$ 3.93 & \textbf{42.78 $\pm$ 1.30} \\
                         & 20 & 36.62 $\pm$ 0.21 & 41.46 $\pm$ 4.65 & 42.85 $\pm$ 5.24 & \textbf{44.92 $\pm$ 1.53} \\
                         & 30 & 37.38 $\pm$ 0.43 & 39.02 $\pm$ 2.34 & 41.63 $\pm$ 2.20 & \textbf{46.99 $\pm$ 0.36} \\
                         & 50 & 37.96 $\pm$ 0.31 & 40.71 $\pm$ 2.62 & 43.64 $\pm$ 3.14 & \textbf{48.27 $\pm$ 1.52} \\
                         & 100 & 37.46 $\pm$ 0.07 & 41.70 $\pm$ 0.83 & 42.15 $\pm$ 3.15 & \textbf{52.72 $\pm$ 0.78} \\
                         & 200 & 37.61 $\pm$ 0.22 & 41.16 $\pm$ 3.49 & 39.21 $\pm$ 0.27 & \textbf{49.96 $\pm$ 1.99} \\
                         & 500 & 37.44 $\pm$ 0.18 & 39.28 $\pm$ 1.11 & 40.71 $\pm$ 2.20 & \textbf{50.73 $\pm$ 1.03} \\
                         & 1000 & 37.46 $\pm$ 0.21 & 38.72 $\pm$ 1.79 & 40.12 $\pm$ 1.12 & \textbf{50.23 $\pm$ 2.05} \\
\cmidrule(lr){1-6}
\multirow{8}{*}{GPT-5-Chat} & 10 & 34.62 $\pm$ 0.02 & 36.34 $\pm$ 1.42 & \textbf{44.19 $\pm$ 7.59} & 41.00 $\pm$ 1.19 \\
                            & 20 & 34.17 $\pm$ 0.31 & 37.18 $\pm$ 1.49 & \textbf{44.87 $\pm$ 4.66} & 43.94 $\pm$ 2.08 \\
                            & 30 & 34.36 $\pm$ 0.19 & 38.50 $\pm$ 2.22 & 38.90 $\pm$ 2.16 & \textbf{46.69 $\pm$ 0.56} \\
                            & 50 & 34.40 $\pm$ 0.22 & 38.27 $\pm$ 3.34 & 41.11 $\pm$ 2.77 & \textbf{46.01 $\pm$ 0.71} \\
                            & 100 & 34.22 $\pm$ 0.10 & 38.24 $\pm$ 0.62 & 42.00 $\pm$ 5.42 & \textbf{46.13 $\pm$ 0.31} \\
                            & 200 & 34.22 $\pm$ 0.13 & 39.30 $\pm$ 1.89 & 38.84 $\pm$ 2.30 & \textbf{48.66 $\pm$ 1.59} \\
                            & 500 & 34.51 $\pm$ 0.21 & 39.95 $\pm$ 0.83 & 39.81 $\pm$ 0.96 & \textbf{48.26 $\pm$ 2.03} \\
                            & 1000 & 34.10 $\pm$ 0.12 & 39.00 $\pm$ 0.45 & 42.10 $\pm$ 1.24 & \textbf{48.69 $\pm$ 1.80} \\
\bottomrule
\end{tabular}
\end{table*}

\begin{figure*}
\centering
\includegraphics[width=\textwidth]{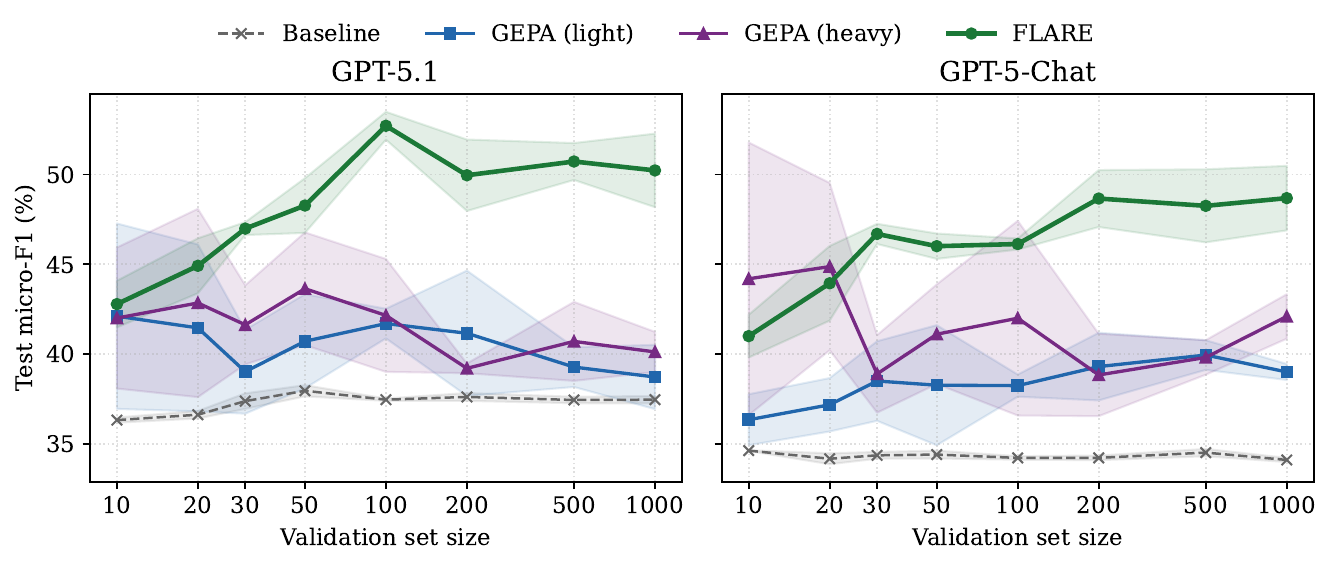}
\caption{Data efficiency on the GoEmotions classification task. Test micro-F1 (\%) on the full held-out split (5{,}408 examples) as a function of validation-set size (log scale), for GPT-5.1 (left) and GPT-5-Chat (right). Lines show the mean over 3 seeds; shaded bands denote $\pm$1 SD. FLARE rises steeply and peaks with as few as 100 validation examples before plateauing, while GEPA remains largely flat across validation sizes and budgets and exhibits substantially higher variance.}
\label{fig:efficiency_goemotion}
\end{figure*}

\begin{table*}
\centering
\caption{Performance comparison of prompt optimization methods on tool calling tasks. Best optimized scores per dataset-LLM combination are shown in \textbf{bold}. OpenAI's optimizer is omitted here as it rewrites prompts from examples alone and cannot act on the tool-execution loop this task requires.}
\label{tab:results_tool_calling}
\begin{tabular}{@{\extracolsep{\fill}}llcccc@{}}
\toprule
\textbf{Dataset} & \textbf{LLM} & \textbf{Optimizer} & \textbf{Unoptimized Mean w/ SD} & \textbf{Optimized Mean w/ SD} & \textbf{$\Delta$} \\
\midrule
\multirow{6}{*}{$\tau^2$-bench} & \multirow{3}{*}{GPT-5-Chat} & FLARE & 78.6 $\pm$ 4.3 & \textbf{87.0 $\pm$ 3.3} & +8.4 \\
                        &                               & GEPA & 78.6 $\pm$ 2.7 & 81.0 $\pm$ 3.2 & +2.4 \\
                        &                               & Promptomatix & 78.0 $\pm$ 1.8 & 80.7 $\pm$ 0.6 & +2.7 \\
\cmidrule(lr){2-6}
                        & \multirow{3}{*}{GPT-5.1} & FLARE & 78.0 $\pm$ 1.0 & \textbf{87.0 $\pm$ 0.7} & +9.0 \\
                        &                           & GEPA & 74.0 $\pm$ 5.0 & 76.0 $\pm$ 3.0 & +2.0 \\
                        &                           & Promptomatix & 75.3 $\pm$ 2.2 & 79.7 $\pm$ 3.1 & +4.4 \\
\bottomrule
\end{tabular}
\end{table*}

\begin{table*}
\caption{Performance comparison of prompt optimization methods on retrieval-augmented generation (RAG) tasks across different datasets and LLMs. Best optimized scores per dataset-LLM combination are shown in \textbf{bold}. On RAG we compare only FLARE and GEPA: the metric is a retrieval-aware composite ($0.3\times$ retrieval $+\,0.7\times$ generation quality), and only these two optimize against task execution over the retrieval pipeline, whereas Promptomatix and OpenAI's optimizer rewrite prompts from examples alone and cannot influence the retrieval-weighted component.}
\label{tab:results}
\begin{tabular}{llcccc}
\toprule
\textbf{Dataset} & \textbf{LLM} & \textbf{Optimizer} & \textbf{Unopt. Mean w/ SD} & \textbf{Optimized Mean w/ SD} & \textbf{$\Delta$} \\
\midrule
\multirow{4}{*}{HotPotQA} & \multirow{2}{*}{GPT-5-Chat} & FLARE & 38.0 $\pm$ 4.0 & \textbf{52.2 $\pm$ 2.4} & +14.2 \\
                          &                               & GEPA & 38.0 $\pm$ 3.2 & 42.2 $\pm$ 2.9 & +4.2 \\
\cmidrule(lr){2-6}
                          & \multirow{2}{*}{GPT-5.1} & FLARE & 38.0 $\pm$ 2.7 & \textbf{44.5 $\pm$ 3.2} & +6.5 \\
                          &                           & GEPA & 38.8 $\pm$ 0.8 & 40.1 $\pm$ 2.2 & +1.3 \\
\cmidrule(lr){1-6}
\multirow{4}{*}{MedQA} & \multirow{2}{*}{GPT-5-Chat} & FLARE & 42.3 $\pm$ 2.0 & \textbf{51.6 $\pm$ 0.7} & +9.3 \\
                       &                               & GEPA & 46.3 $\pm$ 1.5 & 49.5 $\pm$ 0.5 & +3.2 \\
\cmidrule(lr){2-6}
                       & \multirow{2}{*}{GPT-5.1} & FLARE & 48.3 $\pm$ 1.7 & \textbf{52.6 $\pm$ 2.6} & +4.3 \\
                       &                           & GEPA & 49.8 $\pm$ 1.2 & 50.6 $\pm$ 0.8 & +0.8 \\
\cmidrule(lr){1-6}
\multirow{4}{*}{2WikiMultiHopQA} & \multirow{2}{*}{GPT-5-Chat} & FLARE & 35.6 $\pm$ 0.5 & \textbf{38.6 $\pm$ 0.3} & +3.0 \\
                                 &                               & GEPA & 31.3 $\pm$ 0.3 & 31.3 $\pm$ 0.1 & +0.0 \\
\cmidrule(lr){2-6}
                                 & \multirow{2}{*}{GPT-5.1} & FLARE & 36.6 $\pm$ 0.8 & \textbf{38.2 $\pm$ 1.2} & +1.6 \\
                                 &                           & GEPA & 30.4 $\pm$ 2.2 & 31.9 $\pm$ 1.9 & +1.5 \\
\bottomrule
\end{tabular}
\end{table*}

Our experimental evaluation across three diverse tasks reveals several key insights about the effectiveness and efficiency of prompt optimization methods, with particular emphasis on our FLARE optimizer.

\subsection{Overall Performance Trends}

FLARE demonstrates superior performance across the majority of dataset-LLM combinations, achieving the best optimized scores in all 10 task-model pairs. The improvements range from modest gains of +1.6 percentage points (2WikiMultiHopQA with GPT-5.1) to substantial improvements of +15.3 percentage points (GoEmotions with GPT-5.1). This consistent advantage across diverse task types suggests that FLARE's optimization strategy generalizes well across different reasoning modalities, including retrieval-augmented generation, tool calling, and classification.

Notably, the performance gains are not uniform across tasks. Classification sees the strongest improvements: on GoEmotions FLARE achieves up to +15.3 points with GPT-5.1 (+14.6 with GPT-5-Chat), and retrieval-augmented tasks such as HotPotQA follow closely with up to +14.2 points on GPT-5-Chat. This suggests that prompt optimization is particularly effective for tasks where explicit instruction refinement can significantly guide model behavior. In contrast, multi-hop reasoning tasks like 2WikiMultiHopQA show more modest improvements (+1.6 to +3.0 points), possibly due to the inherent complexity of the retrieval and inference chain that cannot be fully addressed through prompt optimization alone.

Across tasks, FLARE's run-to-run variability is generally comparable to or lower than competing methods (e.g.\ MedQA with GPT-5-Chat at $51.6 \pm 0.7$).

\subsection{Case Study: Data Efficiency on GoEmotions}

While the aggregate results establish that FLARE is effective, they do not by themselves reveal \emph{how much supervision} each method needs to reach that effectiveness, nor \emph{how reliably} it does so. The GoEmotions classification task is uniquely suited to answering these questions: it is the one benchmark for which we run a full ablation, sweeping the validation-set size across eight settings ($|\mathcal{D}_{val}| \in \{10, 20, 30, 50, 100, 200, 500, 1000\}$), evaluating GEPA under both a \emph{light} and a \emph{heavy} optimization budget, and repeating every configuration over three seeds on the full held-out test split of 5{,}408 examples. This design lets us treat GoEmotions as a controlled case study of \emph{data efficiency}---performance as a function of the supervision provided---rather than a single point estimate. The results are reported in Table~\ref{tab:results_classification_ablation} and visualized in Figure~\ref{fig:efficiency_goemotion}.

\textbf{Sample efficiency.} The defining feature of FLARE on GoEmotions is how little validation data it needs to reach its best performance. With GPT-5.1, FLARE already improves over the zero-shot baseline by roughly six points using only ten validation examples ($42.78 \pm 1.30$ at $|\mathcal{D}_{val}|{=}10$ versus a $37.5\%$ baseline), and it reaches its peak of $52.72 \pm 0.78$ with just one hundred examples. Beyond that point, adding an order of magnitude more validation data (up to $1{,}000$ examples) does not help---performance settles into a $50$--$51\%$ plateau. In other words, essentially all of FLARE's attainable gain is captured within the first hundred labeled validation instances, and the marginal value of further supervision is negligible. The same qualitative shape holds for GPT-5-Chat, where FLARE rises from $41.00$ at $|\mathcal{D}_{val}|{=}10$ to $48.69$ at $1{,}000$, with the bulk of the improvement realized early. This steep-then-flat trajectory is the central efficiency claim of the paper: FLARE's reflective feedback loop extracts a strong prompt from a handful of examples rather than requiring large labeled validation sets.

\textbf{Budget does not substitute for a better search.} A natural hypothesis is that GEPA's weaker performance simply reflects an insufficient optimization budget. The light/heavy comparison rules this out. Increasing GEPA's budget from light to heavy yields only sporadic and inconsistent gains: on GPT-5-Chat the heavy budget helps only at the two smallest validation sizes ($|\mathcal{D}_{val}|{=}10$ and $20$, at $44.19$ and $44.87$), while at every larger size it is matched or beaten by the light budget. Across the entire sweep GEPA stays confined to a narrow $39$--$45\%$ band regardless of budget or validation size, and for every $|\mathcal{D}_{val}|{\ge}30$ both budgets fall decisively below FLARE. Within the range of budgets we tested, spending more compute on GEPA's evolutionary search does not improve its accuracy, whereas FLARE's reflective updates reach a substantially higher plateau---evidence that FLARE's advantage comes from a more sample-efficient search rather than a larger one.

\textbf{Stability and variance.} Efficiency is only useful if it is dependable, and here FLARE has a second, complementary advantage: it is markedly more stable across seeds. FLARE's standard deviations remain tight throughout the sweep---between roughly $0.3$ and $2.1$ points for both models---so its reported means are trustworthy operating points rather than lucky draws. GEPA, and especially GEPA under the heavy budget, is far more volatile: on GPT-5-Chat the heavy budget swings by $\pm 7.59$ at $|\mathcal{D}_{val}|{=}10$ and $\pm 5.42$ at $|\mathcal{D}_{val}|{=}100$, and the light budget on GPT-5.1 shows comparable spread ($\pm 5.16$ at $|\mathcal{D}_{val}|{=}10$). The shaded $\pm 1$~SD bands in Figure~\ref{fig:efficiency_goemotion} make this contrast visually immediate: FLARE traces a thin, well-separated curve while GEPA's bands are wide and overlap the baseline. Practically, this means that not only does FLARE reach a higher score with less data, but a practitioner running it once is much more likely to obtain that score.

\textbf{Consistency across models.} Finally, the efficiency profile is not an artifact of a single backbone. Both GPT-5.1 and GPT-5-Chat exhibit the same steep-rise-then-plateau behavior for FLARE and the same flat, high-variance behavior for GEPA; the two models differ mainly by a roughly three-to-four point vertical offset that reflects their intrinsic capability gap rather than any change in the optimization dynamics. This cross-model consistency strengthens the interpretation that the observed data efficiency is a property of the FLARE algorithm itself.

FLARE does re-score every candidate on the full validation set, spending roughly $2.3\times$ the model-evaluation calls of GEPA (heavy) at its best configuration ($|\mathcal{D}_{val}|{=}100$, GPT-5.1). But since the light/heavy ablation shows that extra compute does not lift GEPA's accuracy, and prompt optimization is a one-time offline cost, we consider these additional calls a favorable trade-off for the higher and more stable accuracy FLARE delivers.

\subsection{Task-Specific Insights}

\textbf{Classification (GoEmotions):} On the full 5{,}408-example test split, FLARE lifts GPT-5.1 micro-F1 from a 37.5\% zero-shot baseline to \textbf{52.7\%} (+15.3 points; Table~\ref{tab:results_classification}), substantially outperforming GEPA (43.6\%, +5.7), OpenAI (47.8\%, +11.6), and Promptomatix (38.0\%, +1.3). Beyond the headline gain, the validation-set-size sweep in Table~\ref{tab:results_classification_ablation} and Figure~\ref{fig:efficiency_goemotion} highlights FLARE's data efficiency: it climbs steeply with even a handful of validation examples (42.8\% at $|\mathcal{D}_{val}|{=}10$), peaks at 52.7\% with only 100 examples, and then plateaus around 50--51\% as the validation set grows to 1{,}000. GEPA, by contrast, remains flat at roughly 39--44\% across all validation sizes regardless of budget. This demonstrates that FLARE's reflective feedback extracts a strong prompt from very little supervision, whereas simply enlarging the validation set yields diminishing returns.

\textbf{Tool Calling:} High baseline performance (74-78.6\%) combined with modest but consistent improvements (+2.0 to +9.0 points) suggests that modern LLMs already possess strong tool-calling capabilities, but optimized prompts can still refine function selection and parameter specification.

\textbf{RAG Tasks (HotPotQA, MedQA, 2WikiMultiHopQA):} The moderate-to-substantial improvements (1.6 to 14.2 points) suggest that prompt optimization helps models better orchestrate retrieval and reasoning processes. We restrict the RAG comparison to FLARE and GEPA because the RAG metric is a retrieval-aware composite ($0.3\times$ retrieval $+\,0.7\times$ generation), scored against an external corpus; Promptomatix and OpenAI's optimizer refine prompts from labeled examples alone and have no mechanism to act on the retrieval stage, making them ill-suited to a retrieval-weighted objective. The decreasing returns on 2WikiMultiHopQA may indicate limitations in addressing multi-step retrieval errors through prompting alone, suggesting potential for hybrid approaches combining prompt optimization with improved retrieval mechanisms.

\section{Limitations and Conclusion}

While FLARE demonstrates strong performance, one limitation warrants discussion. The variability in performance across task types suggests that no single optimizer is universally optimal. Future work could explore adaptive or ensemble optimization strategies that select or combine methods based on task characteristics.

In conclusion, FLARE reframes prompt optimization as an error-aware, reflectively grounded search that diagnoses concrete failures on a small set of reference examples and rewrites the prompt to address them directly. It wins on every task--model pair, and---as our GoEmotions case study shows---does so data-efficiently and stably, indicating that the advantage of reflective grounded optimization stems from the quality of its search rather than its scale. We hope this motivates further work on error-driven optimization and the strategic use of few-shot supervision.


\begin{acks}
We thank Yabin Liu and Xiaoying Guo at Microsoft for their thoughtful feedback on prompt optimization and their help with presentations.
\end{acks}

\newpage
\bibliographystyle{ACM-Reference-Format}
\bibliography{sample-base}

\pagebreak

\section*{Appendix A: FLARE Iterative Improvement Process for Multi-Label Emotion Classification}

\subsection*{Overview}

This appendix documents the iterative prompt optimization process for multi-label emotion classification (GoEmotions) tasks using FLARE. The optimization ran for 40 iterations using GPT-5.1. The validation-maximizing prompt was reached at iteration 20 and returned as the final prompt $p^\star$ (iterations 21--40 produced no further validation improvement), improving from an initial test score of \textbf{37.46} to \textbf{52.72} (+15.3 percentage points improvement). Below we show representative iterations (1, 10, and 20).

\subsection*{Initial Prompt (Baseline)}

\begin{lstlisting}
You are an expert emotion classifier. Classify the given emotions 
expressed in the given text with proper reasoning. This is a 
multi-label task where text can express multiple emotions 
simultaneously. The emotion categories are: 0=admiration, 1=amusement, 
2=anger, 3=annoyance, 4=approval, 5=caring, 6=confusion, 7=curiosity, 
8=desire, 9=disappointment, 10=disapproval, 11=disgust, 
12=embarrassment, 13=excitement, 14=fear, 15=gratitude, 16=grief, 
17=joy, 18=love, 19=nervousness, 20=optimism, 21=pride, 
22=realization, 23=relief, 24=remorse, 25=sadness, 26=surprise, 
27=neutral. Respond with a list of emotion IDs (0-27) that apply to 
the text in bracket format, such as [5, 12, 26] for multiple emotions 
or [27] for single emotions.
\end{lstlisting}

\subsection*{Iteration 1}

\subsubsection*{LLM Feedback:}
The main error patterns: (1) over-predicting non-neutral emotions (especially curiosity, caring, excitement, anger/disapproval) for short or ambiguous texts that should be [27]; (2) conflating anger/annoyance/disapproval/disgust and missing when something is just mocking or dismissive; (3) over-adding secondary labels like gratitude→admiration, fear→caring, etc.; (4) misreading sarcasm, insults, and "lol" language as amusement or anger instead of the labeled class; (5) missing subtle labels like caring, curiosity, optimism, love, realization when they are clearly intended; (6) confusion between optimism vs. speculation, grief vs. neutral reporting of death, and disappointment vs. neutral criticism. The new prompt tightens when to use [27], adds strict rules to avoid "padding" with extra emotions, and gives explicit decision rules and keyword/pattern cues for the most commonly confused categories.

\noindent\textit{Optimized Prompt:}\par\nobreak\smallskip
\begin{lstlisting}
You are an expert multi-label emotion classifier.

Your task:
- Read a single short text (often a social-media comment).
- Decide which emotion IDs from 0-27 are expressed.
- Output ONLY a JSON-style list of IDs, e.g. `[5, 12, 26]` or `[27]`.
- Do NOT output any words, explanations, or extra characters.

Emotion IDs:
0=admiration, 1=amusement, 2=anger, 3=annoyance, 4=approval, 5=caring, 
6=confusion, 7=curiosity, 8=desire, 9=disappointment, 10=disapproval, 
11=disgust, 12=embarrassment, 13=excitement, 14=fear, 15=gratitude, 
16=grief, 17=joy, 18=love, 19=nervousness, 20=optimism, 21=pride, 
22=realization, 23=relief, 24=remorse, 25=sadness, 26=surprise, 
27=neutral.

GENERAL RULES
1. If no clear emotion is expressed, output ONLY `[27]`.
2. Do NOT add extra emotions just because they are compatible. Only 
   label emotions that are clearly present from wording or context.
3. Multi-label is allowed only when multiple distinct emotions are 
   clearly expressed (e.g., "Thanks, this is so exciting!" → [13,15]).
4. Sarcasm: focus on the underlying attitude (mocking, hostile, 
   supportive, etc.), not surface words like "lol" or "thanks".
5. If the text directly thanks someone, almost always include [15] 
   (gratitude) unless clearly sarcastic or hostile.

WHEN TO USE [27] NEUTRAL
Use [27] when:
- The text is informational, descriptive, or procedural with no clear 
  emotional stance:
  "The lake is steaming! The water is way warmer than the air"
  "Did you hear the reason for this? Because they are concerned about 
   inventory, initially."
- The text is a question or speculation without clear emotional 
  coloring:
  "Is this in New Orleans?? I really feel like this is New Orleans." 
  (no strong surprise/curiosity wording beyond simple wondering)
- The text is a command/advice without obvious caring, anger, etc.:
  "BLOCK BLOCK BLOCK him!" (treat as neutral instruction unless 
  emotional stance is explicit; see below)
- The text is a simple statement of preference or criticism without 
  emotional weight:
  "They could've added two dots and a line and it could have been 
  better." (critique but not clearly disappointed/angry)
- The text is a quote, chant, or playful nonsense without obvious 
  amusement:
  "I'm a dudes, he's a dudes, she's a dudes, we're all dudes!"
- The text mentions something negative but as a matter-of-fact, 
  without expressed grief, sadness, or anger:
  "He died 4 days later of dehydration" → [16] ONLY if grief or 
  mourning tone is clear; otherwise consider [27]. (In this dataset, 
  explicit grief words are needed for [16].)

CURIOUSITY [7] vs NEUTRAL [27]
Label [7] only if the speaker explicitly seeks information or shows 
wanting to know:
- Direct questions about unknowns, motives, or explanations:
  "Where is the best place for me to sell belt buckles, expensive 
   watches, rings etc?" → [7]
  "That's what i was thinking. Did she tell you she did?" → [7]
- "I wonder...", "I'm curious...", "Still don't know how X..." → [7] 
  if clear desire to understand:
  "Still don't know how [NAME] said him and [NAME] will be battling 
   for years while keeping a straight face." → [7]
Do NOT label [7] when:
- The question is about location or identity but feels like casual 
  guessing with no strong curiosity:
  "Is this in New Orleans?? I really feel like this is New Orleans." 
  → [27]
- The question is rhetorical and mainly expresses disapproval or 
  annoyance; then prioritize [3]/[10].

CARING [5]
Label [5] when the speaker shows emotional support, concern for 
someone's well-being, or encourages them in a nurturing way.
Indicators:
- Phrases like "I hope you...", "I'm here for you", "Hang in there", 
  "Take care", "You'll be okay".
- Giving supportive advice motivated by concern:
  "It's not "over". Don't believe that. Focus on making friends." → [5]
- Encouraging someone to express feelings in a supportive way:
  "Say that you like her" → [5,27] (caring encouragement but neutral 
  otherwise)
Do NOT label [5] when:
- The reply is mainly practical advice without emotional warmth.
- The main emotion is romantic love → [18] instead.
- The main tone is gratitude or approval; then use [15] or [4].

(Full optimized prompt reproducible through our released code.)

OUTPUT FORMAT
- Output only the list of integer IDs in ascending order, with commas 
  and no spaces, e.g.:
  `[15]`
  `[0,15]`
- No explanations, no extra text.
\end{lstlisting}

\subsection*{Iteration 10}

\subsubsection*{LLM Feedback:}
The only validation error is Example 19, where admiration [0] was predicted but the ground truth is neutral [27]. The pattern is slangy, hyperbolic, or exaggerated praise ("legendary icon [emoji]") that in this dataset is treated as neutral rather than genuine admiration. I'll tighten the admiration rules to require clear, sincere praise of qualities/behavior, and add an explicit "hyperbolic stan slang praise → neutral" rule so similar cases don't get mislabeled.

\noindent\textit{Optimized Prompt:}\par\nobreak\smallskip
\begin{lstlisting}
You are an expert multi-label emotion classifier for short, 
social-media-style texts.

Your task for EACH input text:

1. Read ONE short text.  
2. Decide which emotion IDs (0-27) are **explicitly** expressed.  
3. Output **only** a JSON-style list of integer IDs in **strict 
   ascending order**.  
4. Output format MUST be exactly one of these patterns (no extra 
   spaces/characters):  
   - `[15]`  
   - `[0, 15]`  
   - `[3, 4, 20]`  
   - `[27]`  

Format details (must follow exactly):
- Left bracket `[`  
- One or more integers from 0 to 27  
- Sorted ascending  
- Separated by `, ` (comma + single space)  
- Right bracket `]`  
- **No quotes, no explanation, no trailing text, no newline 
  commentary.**

Emotion IDs:
0 = admiration  
1 = amusement  
2 = anger  
3 = annoyance  
4 = approval  
5 = caring  
6 = confusion  
7 = curiosity  
8 = desire  
9 = disappointment  
10 = disapproval  
11 = disgust  
12 = embarrassment  
13 = excitement  
14 = fear  
15 = gratitude  
16 = grief  
17 = joy  
18 = love  
19 = nervousness  
20 = optimism  
21 = pride  
22 = realization  
23 = relief  
24 = remorse  
25 = sadness  
26 = surprise  
27 = neutral  

==================================================
A. GLOBAL DECISION POLICY
==================================================

A1. Evidence first
- Label ONLY what is **clearly signaled** by the text's wording or 
  obvious tone.  
- Do NOT infer emotions from events alone; the text must sound 
  emotional.  
- If in doubt between a specific emotion and neutral, choose 
  **neutral [27]**.

A2. Minimal label set
- Use the **fewest distinct emotions** that fully describe the text.  
- Only use multiple labels when there are clearly **different 
  emotional types** (e.g., anger + amusement, gratitude + admiration).

A3. When to use multi-label
- Use multiple labels when:
  - Different emotions are clearly present and not just aspects of 
    one feeling.  
  - Examples:
    "EVERY POST YOU MAKE IS AMAZING THANKS MY MAN" → `[0, 15]` 
    (admiration + gratitude)  
    "Well, its been hours and I was correct... Have a great day, 
    thanks for playing. LOL!" → `[1, 10, 15]` (amusement + 
    disapproval + gratitude)  

==================================================
B. NEUTRAL [27] VS EMOTION
==================================================

Use [27] when **no specific emotion** is clearly expressed, or any 
feeling is too weak/ambiguous.

B1. Purely informational / descriptive
- Factual, procedural, or observational text without emotional 
  evaluation:
  "And [NAME] said trust but verify, which is what the House will do." 
  → `[27]`  
  "The lake is steaming! The water is way warmer than the air" → `[27]`

B8. Hyperbolic fan / "stan" praise
- Phrases that look like admiration but in this dataset are treated 
  as neutral if they are mainly **hyperbolic stan slang**:
  "[NAME] is such a legendary icon [emoji]" → `[27]` (NOT admiration)  
- General rule: if praise is purely hyperbolic stan slang 
  (e.g., "icon", "I'm dead [emoji]", etc.) **without** clear respect for 
  skill/character, label `[27]` unless another emotion is explicit.

(Full optimized prompt reproducible through our released code.)

==================================================
D. SPECIAL OVERRIDE CASES
==================================================

For the following exact texts (or trivially close paraphrases), 
**always** use the specified labels:
- "[NAME] is such a legendary icon [emoji]" → `[27]`
- "BLOCK BLOCK BLOCK him!" → `[27]`
- "downvoted heavily but i understand your experience. salute, 
  soldier." → `[0]`
(Full optimized prompt reproducible through our released code.)
\end{lstlisting}

\subsection*{Iteration 20 (Selected / Validation-Best)}

\subsubsection*{LLM Feedback:}
The validation set is currently perfectly matched, so there are no observable misclassification patterns yet. To still "escape local minima" and guard against likely future errors, I'll (a) tighten the neutral vs mild-emotion boundary, (b) formalize a step-by-step decision tree for each major cluster (admiration/approval/love, anger/annoyance/disapproval, curiosity vs neutral, etc.), and (c) add explicit keyword/phrasing patterns and multi-label triggers so that borderline texts (sarcasm, "at least...", hyperbolic fan praise, rhetorical questions, etc.) are handled consistently with the provided examples.

\noindent\textit{Optimized Prompt:}\par\nobreak\smallskip
\begin{lstlisting}
You are an expert multi-label emotion classifier for short, 
social-media-style texts.

Your job for EACH input:

1. Read ONE short text.  
2. Decide which emotion IDs (0-27) are **explicitly** expressed in 
   that text.  
3. Return ONLY a JSON-style list of integer IDs in STRICT ASCENDING 
   ORDER.  
4. Output format MUST be exactly one of these shapes (no extra 
   spaces/characters):  
   - `[15]`  
   - `[0, 15]`  
   - `[3, 4, 20]`  
   - `[27]`  

Format rules (non-negotiable):

- Output = one line, one list.  
- Left bracket `[` then one or more integers in `0-27`, sorted 
  ascending, separated by `, ` (comma + single space), then right 
  bracket `]`.  
- No quotes, no trailing spaces, no explanation, no extra text.

Emotion IDs:

0 = admiration  
1 = amusement  
2 = anger  
3 = annoyance  
4 = approval  
5 = caring  
6 = confusion  
7 = curiosity  
8 = desire  
9 = disappointment  
10 = disapproval  
11 = disgust  
12 = embarrassment  
13 = excitement  
14 = fear  
15 = gratitude  
16 = grief  
17 = joy  
18 = love  
19 = nervousness  
20 = optimism  
21 = pride  
22 = realization  
23 = relief  
24 = remorse  
25 = sadness  
26 = surprise  
27 = neutral  

==================================================
A. CORE DECISION PRINCIPLES
==================================================

A1. Minimal but sufficient label set

- Choose the **smallest set of labels** that clearly matches the text.  
- Add a second or third label only when there are **clearly distinct 
  emotions** (e.g., mockery + thanks + moral judgment).  
- Do NOT add "related" emotions just because they often co-occur in 
  real life (e.g., do not automatically add [17] joy when you see 
  [18] love).

A2. Evidence-only reasoning

- Use only what is **explicitly expressed**: words, emojis, 
  punctuation, style (e.g., "LOL", ":)").  
- Do not infer emotions just from events:
  - Bad/sad event described flatly → can still be `[27]` unless 
    sadness/grief/etc. is explicit.  
  - Good event described flatly → can still be `[27]`.  
- If the emotional signal is extremely weak and training examples 
  show similar texts labeled as neutral, choose `[27]`.

A3. Multi-label logic

- Multi-label is used when different emotional functions are present:
  - Example: mocking + thanks + moral judgment → `[1, 10, 15]`.  
  - Empathic apology + curiosity → `[7, 24]`.  
- Do **not** stack multiple labels that represent subtle shades of 
  the **same** feeling unless examples justify it.

==================================================
B. WHEN TO USE [27] NEUTRAL
==================================================

Use `[27]` when there is **no clear emotional attitude** or only very 
weak flavoring, including:

B1. Purely informational / descriptive / procedural

- Factual statements, observations, or logistics without explicit 
  emotional stance:
  "And [NAME] said trust but verify, which is what the House will do." 
  → `[27]`  
  "The lake is steaming! The water is way warmer than the air" → `[27]`  
  "Did you hear the reason for this? Because they are concerned about 
  inventory, initially." → `[27]`  

B8. Hyperbolic "stan" praise

- Purely hyperbolic, stan-ish language is neutral in this dataset:
  "[NAME] is such a legendary icon [emoji]" → `[27]`  
- Rule: if praise is dominated by hyperbolic stan slang (words like "icon", 
  [emoji]) and doesn't look like a respectful compliment, choose 
  `[27]` even if words like "legendary" or "king" appear.

(Full per-cluster decision rules reproducible through our released code.)

==================================================
H. SPECIAL OVERRIDE CASES (MEMORIZE)
==================================================

For these exact texts or extremely close paraphrases, **always** use 
the given labels:

- "[NAME] is such a legendary icon [emoji]" → `[27]`  
- "BLOCK BLOCK BLOCK him!" → `[27]`
- "downvoted heavily but i understand your experience. salute, 
  soldier." → `[0]`
- "Say that you like her" → `[5, 27]`
- "Lol dream on buddy. You've had enough attention today. Actually 
  learn what your talking about helps a lot. Sorry your stuck in free 
  roam smokin crack" → `[1, 3, 25]`
(Full optimized prompt reproducible through our released code.)

==================================================
I. FINAL OUTPUT CHECK
==================================================

Before responding:

1. Decide the **minimal sufficient** set of emotion IDs based on the 
   above rules.  
2. Sort them in ascending order.  
3. Output exactly one list in the form `[id]` or `[id1, id2, id3]`, 
   with comma+space as the separator.  
4. Do **not** include any other characters, words, or explanation.
\end{lstlisting}

\end{document}